\documentclass[a4paper]{article}

\usepackage[english]{babel}
\usepackage[utf8x]{inputenc}
\usepackage[T1]{fontenc}
\usepackage{authblk}
\usepackage{algorithm}
\usepackage{algpseudocode}
\usepackage{pgfplots}

\usepackage[a4paper,top=3cm,bottom=2cm,left=3cm,right=3cm,marginparwidth=1.75cm]{geometry}

\usepackage{amsmath}
\usepackage{graphicx}
\usepackage[colorinlistoftodos]{todonotes}
\usepackage[colorlinks=true, allcolors=blue]{hyperref}
\usepackage{booktabs}

\title{XGBoost: Scalable GPU Accelerated Learning}
\author[1]{Rory Mitchell}
\author[2]{Andrey Adinets}
\author[3]{Thejaswi Rao}
\author[4]{Eibe Frank}
\affil[1,4]{University of Waikato}
\affil[1]{H2O.ai}
\affil[2,3]{Nvidia Corporation}
\affil[*]{Corresponding author: Rory Mitchell, ramitchellnz@gmail.com, Waikato University Computer Science Department}

\begin{document}
\maketitle

\begin{abstract}
We describe the multi-GPU gradient boosting algorithm implemented in the XGBoost library\footnote{https://github.com/dmlc/xgboost}. Our algorithm allows fast, scalable training on multi-GPU systems with all of the features of the XGBoost library. We employ data compression techniques to minimise the usage of scarce GPU memory while still allowing highly efficient implementation. Using our algorithm we show that it is possible to process 115 million training instances in under three minutes on a publicly available cloud computing instance. The algorithm is implemented using end-to-end GPU parallelism, with prediction, gradient calculation, feature quantisation, decision tree construction and evaluation phases all computed on device. 

\end{abstract}

\providecommand{\keywords}[1]{{\textbf{\textit{Keywords---}} #1}}

\section{Introduction}
Gradient boosting is an algorithmic approach providing state-of-the-art performance on supervised learning tasks such as classification, regression and ranking. XGBoost is an implementation of a generalised gradient boosting algorithm that has become a tool of choice in machine learning applications. This is due to its excellent predictive performance, highly optimised multicore and distributed machine implementation, and the ability to handle sparse data. We describe extensions to this library enabling multi-GPU-based execution, significantly reducing the runtime of large-scale problems. These extensions are advances on the original GPU accelerated gradient boosting algorithm in~\cite{accelerating_xgboost}. This new work presents a significantly faster and more memory efficient decision tree algorithm based on feature quantiles, as well as parallel algorithms for other parts of the gradient boosting algorithm. We implement decision tree construction, quantile generation, prediction, and gradient calculation algorithms all on the GPU to accelerate the gradient boosting pipeline from end to end. This allows the XGBoost library to utilise the significantly enhanced memory bandwidth and massive parallelism of GPU-enabled systems.

Our GPU-accelerated gradient boosting extensions are available through the standard XGBoost API when compiled with GPU support. Hence, they may be used from C++, Python, R, and Java and support all of the standard XGBoost learning tasks such as regression, classification, multiclass classification, and ranking. Both Windows and Linux platforms are supported. Like the original XGBoost algorithm, our implementation fully supports sparse input data.

\section{Description}
\begin{figure}
\caption{Gradient Boosting Pipeline}
\label{fig:pipeline}
\includegraphics[width=\textwidth]{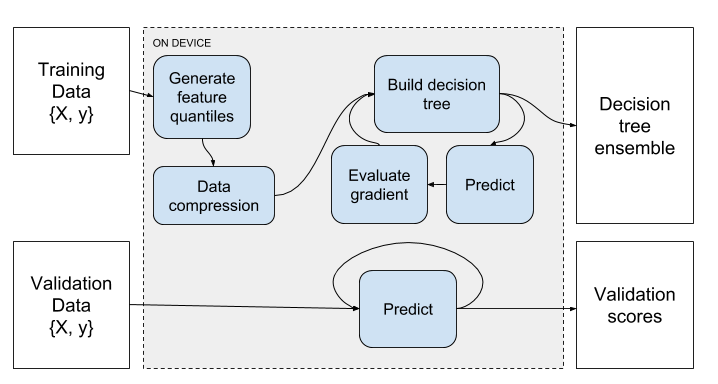}
\end{figure}

Supervised gradient boosting takes a labelled training dataset as input and iteratively defines a series of trees that progressively refine accuracy with respect to an objective function. Figure~\ref{fig:pipeline} shows this process at a high level. We implement all of these significant operations on one or multiple GPUs, described in the following sections.

\subsection{Feature Quantile Generation}
The implementation described in this paper works on a quantile representation of the input feature space, as described in~\cite{Keck2017,Ke2017}. Working with features in quantised form reduces the tree construction problem largely to one gradient summation into histograms, speeding up execution time considerably. Quantising the input matrix is a preprocessing step that involves considerable computation, and so we map it to the GPU. 

\subsection{Data Compression}
We apply a compression step to the quantised matrix to take advantage of the reduced cardinality of the feature space. Matrix values are compressed down to \(log_2(max\_value)\) bits, where \(max\_value\) is the maximum integer value of any quantised matrix element. Values are packed and unpacked at runtime using bitwise operations. Performing this operation at runtime is more flexible than precompiling many versions of the program. The small number of bitwise operations computed on the GPU incur no visible performance penalty. This compression step typically reduces GPU memory consumption by four times or more over the standard floating point representation, allowing training on significantly larger datasets. 

\subsection{Decision Tree Construction}
\algrenewcommand{\algorithmiccomment}[1]{$\triangleright$ #1}
\begin{algorithm}
\caption{Decision Tree Construction}
\label{alg:tree_construction}
\begin{algorithmic}
\State \textbf{Input:} 
\State \(X_1, X_2 ... X_p \in X\): training examples partitioned onto \(p\) GPUs
\State \(\vec{g_1}, \vec{g_2} ... \vec{g_p} \in \vec{g}\): gradients for training examples partitioned onto \(p\) GPUs
\State \textbf{Output:} \(tree\): set of output nodes
\State tree \(\leftarrow\) \{\}
\State expand\_queue \(\leftarrow\) InitRoot()
\While{expand\_queue is not empty}
	\State expand\_entry \(\leftarrow\) expand\_queue.pop()
    \State tree.insert(expand\_entry)
    \State \Comment{Process a subset of training instances on each GPU}
    \For{\(i\) in \(0..p\)}
    	\State \Comment{Sort training instances into leaf nodes based on previous split}
    	\State RepartitionInstances(expand\_entry, \(Xi\))
    	\State \Comment{Build partial gradient histograms}
    	\State BuildPartialHistograms(expand\_entry, \(Xi\), \(\vec{g_i}\))
    \EndFor
    \State \Comment{Combine histograms across all GPUs}
    \State AllReduceHistograms(expand\_entry)
    \State \Comment{Find the optimal split for both the left and right children}
    \State left\_expand\_entry \(\leftarrow\) EvaluateSplit(expand\_entry.left\_histogram)
    \State right\_expand\_entry \(\leftarrow\) EvaluateSplit(expand\_entry.right\_histogram)
    \State expand\_queue.push(left\_expand\_entry)
    \State expand\_queue.push(right\_expand\_entry)
\EndWhile
\end{algorithmic}
\end{algorithm}
We apply the multi-GPU decision tree construction algorithm shown in Algorithm \ref{alg:tree_construction}. Each GPU processes a subset of training instances and is responsible for partitioning these instances into leaf nodes and calculating partial gradient histograms. The partial histograms are merged using an AllReduce operation provided by the NCCL library~\cite{nccl}. The split gain may then be calculated for each feature and each quantile by performing a scan over the gradient histogram. This is achieved on the GPU with a parallel prefix sum operation~\cite{prefixsum}.
The tree growth strategy in this algorithm is reconfigurable to prioritise expanding nodes with a higher reduction in the objective function or nodes closer to the root.

\subsection{Prediction}
The gradient boosting algorithm typically requires making predictions on the training set as an input to gradient calculation at each boosting iteration, as well as predictions on a test or validation to evaluate progress. We map this computation to the device by assigning each GPU thread one training instance and then iterating through each tree sequentially. This operation is a relatively poor fit for the GPU architecture as traversing a tree requires the evaluation of many branch statements. However, we are still able to significantly outperform the CPU implementation, due to much higher memory bandwidth, increased parallelism, and the fact that CPU architectures also deal poorly with large numbers of unpredictable branch statements.

\subsection{Gradient Evaluation}
In XGBoost, at each boosting iteration, the first and second order gradient with respect to the objective function must be calculated for each training instance. Equations \ref{gradient} and \ref{hessian} show the calculation of the gradient and hessian values for logistic loss on training instance \(i\) with label \(y_i\) and predicted margin \(\hat{y_i}\). This computation is trivially mapped to the GPU with one training instance per working thread and results in significant performance improvement compared to a CPU-based implementation due to the larger FLOPS and memory bandwidth of the GPU architecture.

\begin{equation}
\label{gradient}
g_i = sigmoid(\hat{y_i}) - y_i 
\end{equation}
\begin{equation}
\label{hessian}
h_i = sigmoid(\hat{y_i})(1 - sigmoid(\hat{y_i}))
\end{equation}

We support on device gradient calculation for logistic and linear regression objectives. Other objectives such as multiclass and ranking are supported but will be calculated on the CPU. GPU versions of these objectives are a work in progress.

\section{Evaluation}
We evaluate our implementation against its two primary competitors LightGBM~\cite{lightgbm,lightgbm_gpu} and CatBoost\cite{catboost}. Both implement multi-core CPU and GPU training. The evaluation is performed on a cloud instance with 8 Tesla V100 GPUs and 64 CPU cores\footnote{https://aws.amazon.com/ec2/instance-types/p3/}. We benchmark on the 6 publicly available data sets shown in Table~\ref{datasets} using 500 boosting iterations. Training time and accuracy are reported in Table~\ref{evaluation}. See our benchmarking repository\footnote{https://github.com/RAMitchell/GBM-Benchmarks} for minimal instructions to reproduce these benchmarks as well as full details on hyperparameters and preprocessing steps.
\begin{table}[H]
\centering
\caption{Datasets}
\label{datasets}
\begin{tabular}{|l|l|l|l|}
\hline
 Name & Rows & Columns & Task \\ \hline
YearPredictionMSD\cite{msdyear} & 515K & 90 & Regression \\\hline
Synthetic\cite{scikit-learn} & 10M & 100 & Regression \\\hline
higgs\cite{higgs} & 11M & 28 & Classification \\\hline
Cover Type\cite{cover_type} & 581K & 54 & Multiclass classification \\\hline
bosch\cite{bosch} & 1M & 968 & Classification \\\hline
airline\cite{airline} & 115M & 13 & Classification \\
\hline
\end{tabular}
\end{table}
\begin{table}[H]
\centering
\caption{Evaluation }
\label{evaluation}
\tabcolsep=0.11cm
\scalebox{0.70}{
\begin{tabular}{rrrrrrrrrrrrr}
\toprule
{} & \multicolumn{2}{l}{YearPrediction} & \multicolumn{2}{l}{Synthetic} & \multicolumn{2}{l}{Higgs} & \multicolumn{2}{l}{Cover Type} & \multicolumn{2}{l}{Bosch} & \multicolumn{2}{l}{Airline} \\
{} &           Time(s) &     RMSE &   Time(s) &     RMSE &  Time(s) &  Accuracy &    Time(s) &  Accuracy &  Time(s) &  Accuracy &  Time(s) &  Accuracy \\
\midrule
xgb-cpu-hist &           216.71 &  8.8794 &   580.72 &  13.6105 &  509.29 & 74.74 &    3532.26 &  89.20 &  810.36 &  \textbf{99.45} & 1948.26 & 74.94 \\
xgb-gpu-hist &           30.39 &  8.8799 &   43.14 &  13.4606 &  38.41 &  \textbf{74.75} &    \textbf{107.70} &  \textbf{89.34} &  \textbf{27.97} &  99.44 &  \textbf{110.29} &  74.95\\
lightgbm-cpu &           30.82 &  8.8777 &   463.79 &   13.5850 &  330.25 &  74.74 &    186.27 &  89.28 &  162.29 &  99.44 & 916.04 & \textbf{75.05}\\
lightgbm-gpu &           25.39 &  \textbf{8.8777} &   576.67 &   13.5850 &  725.91 &   74.70 &     383.03 &  89.26 &  409.93 &  99.44 & 614.74 & 74.99\\
cat-cpu      &           39.93 &  8.9933 &   426.31 &  9.3870 &  393.21 &  74.06 &    306.17 &  85.14 &  255.72 &  99.44 & 2949.04 & 72.66\\
cat-gpu      &           \textbf{10.15} &  9.0637 &   \textbf{36.66} &  \textbf{9.3805} &  \textbf{30.37 }&  74.08 &        N/A &       N/A &      N/A &       N/A & 303.36 & 72.77\\
\bottomrule
\end{tabular}
}
\end{table}
Some results are missing for "cat-gpu" as it does not support multiclass objectives and an error was encountered on the Bosch dataset.

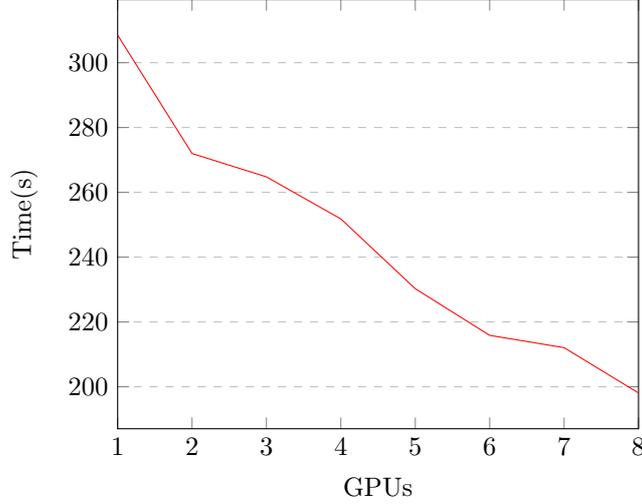
\begin{figure}
\centering
\caption{XGBoost run-time on Airline dataset: 1-8 V100 GPUs}
\label{fig:scalability}
\begin{tikzpicture}
\begin{axis}[
    xlabel={GPUs},
    ylabel={Time(s)},
    xmin=1, xmax=8,
    ymajorgrids=true,
    grid style=dashed,
]
 
\addplot[
    color=red,
    ]
    coordinates {(1,308.504153966903)(2,271.982721646626)(3,264.767537991205)(4,251.783841689427)(5,230.257787307103)(6,215.852016369501)(7,212.066198746363)(8,198.075062274932)
    };
\end{axis}
\end{tikzpicture}
\end{figure}

Our algorithm ("xgb-gpu-hist") is the fastest on 3/6 datasets and the most accurate on 2/6 datasets. In the case of the largest dataset with 115M rows our algorithm is almost 3x faster than its nearest competitor. Its runtime never exceeds two minutes on any dataset.
Figure \ref{fig:scalability} shows runtime with additional GPUs on the airline dataset. After compression and distributing training rows between 8 GPUs, we only require 600MB per GPU to store the entire matrix. 

\section{Conclusion}
Evaluation shows our gradient boosting implementation is highly effective at processing large datasets. The XGBoost GPU algorithm shows high accuracy and low runtime on a wide range of datasets and learning tasks. 

Future work includes extending our implementation to run on distributed machines and handle extremely large files exceeding the capacity of system memory, targeting datasets up to 1TB in size. Using the XGBoost framework, GPU-accelerated algorithms are also being developed for large-scale regularised linear modelling problems.
\bibliographystyle{plain}
\bibliography{references}

\end{document}